\documentclass{article}

\usepackage{latexsym, amssymb, amsfonts, fancyvrb, natbib}

\title{Information Compression by Multiple Alignment, Unification and Search
as a Unifying Principle in Computing and Cognition\protect\footnote{In {\em Artificial Intelligence Review} {\bf 19}(3), 193--230, 2003.}} 

\author{J Gerard Wolff
\\
\\
{\small \it School of Informatics, University of Wales Bangor,}
\\
{\small \it Dean Street, Bangor, LL57 1UT, UK.}
\\
{\small \it E-mail: gerry@informatics.bangor.ac.uk.}
\\
{\small \it Web: www.informatics.bangor.ac.uk/$\sim$gerry/sp\_summary.html.}
}

\normalsize

\begin{document}
\maketitle
\begin{abstract}
\noindent This article presents an overview of the idea that {\em information compression by multiple alignment, unification and search} (ICMAUS) may serve as a unifying principle in computing (including mathematics and logic) and in such aspects of human cognition as the analysis and production of natural language, fuzzy pattern recognition and best-match information retrieval, concept hierarchies with inheritance of attributes, probabilistic reasoning, and unsupervised inductive learning. The ICMAUS concepts are described together with an outline of the SP61 software model in which the ICMAUS concepts are currently realised. A range of examples is presented, illustrated with output from the SP61 model.

\end{abstract}

\section{Introduction}

This article presents the idea that {\em information compression by multiple alignment, unification and search} (ICMAUS) may serve as a unifying principle in artificial computing and natural cognition. In this context, {\em multiple alignment} has a meaning that is similar to its meaning in bio-informatics (with important differences), {\em unification} means a simple merging of patterns or parts of patterns that match each other, and {\em search} means searching for alignments yielding relatively large compression, pruning the search space using heuristic techniques or other forms of constraint.

There is now a fairly large body of evidence in support of this conjecture, described in \citet{wolff_2000, wolff_1999_prob, wolff_1999_comp} and earlier sources cited there. This article presents an overview of these ideas,  with examples from the SP61 computer model. The aim of this paper is to explain the ICMAUS concepts in outline and to illustrate their wide scope. A full evaluation of the concepts is outside the scope of this paper although a few remarks on that subject are made in Section \ref{comparative_evaluation}. Another paper is planned to evaluate the concepts in relation to alternative theories and empirical data.

The emphasis in previous writings about the ICMAUS concepts has been on the `computing' aspects of the proposals, whereas this article puts weight on the multi-disciplinary nature of the proposals and the cognitive science dimension.

\section{Background}\label{background}

\subsection{Cognitive economy}

From the writings of William of Ockham in the 14th century and Ernst Mach in the 19th century, a principle of parsimony has been recognised as relevant to an understanding of thinking, perception and the workings of brains and nervous systems. Other writings in this tradition include \cite{zipf_1949, attneave_1954, oldfield_1954, von_bekesy_1967}, \citet{barlow_1969} and many other publications over a long period), \citet{watanabe_article_1972, garner_1974, wolff_1988, wolff_1993}. Nice reviews of some of the issues and many other pointers to related research are provided by \citet{chater_1996, chater_1999}.

In case these ideas seem a little obscure, consider the phenomenon of recognition. Recognising someone as `the same' from one instant to another or from one occasion to another is essentially a process of merging (and thus compressing) many individual percepts into a single concept. If we wish to treat a given person as a single person, it would be very inconvenient if we did not merge our many perceptions of that person in that way.

In terms of biology, there is every reason to believe that natural selection would have favoured brains and nervous systems with an ability to economise on the storage and transmission of information.\footnote{Different rules may apply to the genetic code itself. DNA is such an economical means of storing information that there may be relatively little selection pressure in favour of economy. And replication of genes may on occasion yield advantages in terms of physiological functions.} Perhaps more importantly, a key function of neural tissue is the inductive prediction of the future from the past and it is known that there is an intimate connection between information compression (IC) and this kind of inductive inference \citep{solomonoff_1964, chater_1996, chater_1999}.

\subsection{Models of language learning and Minimum Length Encoding}

The ideas to be described grew more immediately out of a programme of research developing the MK10 and SNPR computer models of first language learning by children \citep[see][and earlier papers cited there]{wolff_1988, wolff_1982}. A key idea emerging from this research was that many aspects of language learning could be understood in terms of IC or, more precisely, principles of Minimum Length Encoding (MLE)\footnote{MLE is an umbrella term for Minimum Message Length encoding (MML) and Minimum Description Length encoding (MDL).} pioneered by \citet{solomonoff_1964, solomonoff_1986, solomonoff_1997, wallace_boulton_1968, rissanen_1978} and others \citep[see][]{li_vitanyi_1997}.

The key idea in MLE is that, in grammar induction and related kinds of processing, one should seek to minimise $(G + E)$, where $G$ is the size (in bits) of the `grammar' (or comparable structure) under development and $E$ is the size (in bits) of the raw data when it has been encoded in terms of the grammar. This principle guards against the induction of trivially small grammars (where a very small $G$ is offset by a relatively large $E$) and over-large grammars (where $E$ may be small\footnote{If the grammar is very poor it may not even achieve a small $E$.} but this is offset by a relatively large $G$).\footnote{The goal of minimising $G$ and $E$ for a grammar echoes the way science does or should aim for theories with a favourable combination of {\em simplicity} and descriptive or explanatory {\em power}.}

Amongst other things, this idea provides a handle on how it is that a child, apparently without the need for correction by a teacher, negative samples or grading of samples (as postulated by Gold, 1967), can distinguish `correct' generalisations from `incorrect' ones despite the fact that, by definition, both kinds of generalisation have zero frequency in the child's experience \citep{wolff_1988, wolff_1982}.

\subsection{Seeing connections}

The first hint (in my mind) that IC by pattern matching, unification and search might provide a unifying framework for biological  and artificial kinds of computation arose from the following train of thought:

\begin{itemize}

\item A prominent feature of the MK10 and SNPR models of language learning is a process of IC by finding patterns that match each other and then merging (`unifying') multiple instances to make one.\footnote{Provided they are larger than a certain minimum, patterns that match each other represent `redundancy' in information. Hence, redundancy can be reduced (and information can be compressed) by unifying matching patterns. This is the basis of all the simpler standard techniques for information compression.} These models also incorporate a process of searching amongst alternative possible unifications to find those that are relatively good in terms of compression.

\item Pattern matching and search are prominent in systems like Prolog, designed originally for theorem proving. And `unification' in the sense of this article (the merging of matching patterns) is a prominent part of `unification' as it is understood in Prolog and other systems for logical deduction.

\end{itemize}

These and other reflections on the nature of information processing led to the first conjectural proposal that `computing' in some deep sense might be understood as IC by the matching and unification of patterns \citep{wolff_1990}.\footnote{\citet{solomonoff_1986} had already observed that the great majority of problems in science and mathematics may be seen as either `machine inversion' problems or `time limited optimisation' problems, and that both kinds of problems can be solved by inductive inference using MLE principles. Although this is not a proposal that `computing' itself might be understood as IC, the observation does provide some support for that hypothesis.} Given the prominence of IC in diverse aspects of human cognition, perception and neural functioning, it is natural to suppose that IC may provide a unifying principle for both artificial computing and the processing of information in brains and nervous systems \citep{wolff_1993}.

The aim of subsequent research has been to refine and develop these first embryonic ideas, attempting to answer the many questions and issues that arose.  A key insight has been that the explanatory scope of the concepts could be greatly increased by replacing `pattern matching' within the framework by the more specific concept of `multiple alignment' (similar to the concept of multiple alignment in bio-informatics but different in important ways, as will be described).

\subsection{Related research}

Since the ICMAUS concepts are an attempt to integrate ideas across a wide area, they naturally relate to a large volume of published research. As noted above, this article aims to present the ideas and illustrate their wide scope, with the evaluation of the proposals in relation to existing research largely deferred to another paper.

In general terms, this research may be seen as a contribution to Newell's \citeyearpar{newell_1992} quest for `unified theories of cognition', alongside Soar (e.g., \citet{laird_newell_rosenbloom_1987, young_lewis_1999}), ACT-R \citep[e.g.,][]{anderson_lebiere_1998} and other attempts to integrate concepts in cognition. Like these other models, the ICMAUS framework aims to avoid the pitfalls of `micro' theories for narrow domains by developing concepts with broad scope whilst striving for overall simplicity.

The main differences between the ICMAUS proposals and others are:

\begin{itemize}

\item The emphasis on IC as a unifying principle.

\item The multiple alignment concept as it has been developed in this research.

\item The intended scope of the proposals beyond human cognitive psychology to concepts in computing, mathematics and logic.

\end{itemize}

\section{\sloppy The ICMAUS Framework and the SP Models}\label{icmaus_and_sp_models}

The ICMAUS framework is founded on principles of Minimum Length Encoding mentioned above. The framework is intended as an abstract model of any natural or artificial system for computing or cognition.

In broad terms, the framework is an incremental process not unlike the well-known and widely-used LZ algorithms for IC (PkZip, WinZip etc). The system is intended to receive raw (or {\em New}) data from the environment, to compress those data as much as possible and store them in a steadily growing repository of {\em Old} information. Compression of New is achieved by matching New against itself and against patterns already stored in Old, unifying those parts that match (thus achieving the effect of recognition) and storing the parts that do not match (which may be seen as an element of learning).

The main differences between the ICMAUS framework and the LZ algorithms are the provision of mechanisms to achieve a relatively thorough search of the space of alternative unifications, the ability to find good partial matches between patterns, and the ability to create multiple alignments of the kind that will be described.

\subsection{Information can be compressed and redundant at the same time}\label{compressed_and_redundant}

Paradoxical as it may seem, the idea that the workings of computers and brains may be founded on IC is not in conflict with the uses of redundancy in information to reduce errors or speed up processing.

It is widely accepted that databases should minimise redundancy, as far as practicable. If, for example, the name, address etc of a given person appears two or more times in a database, then updating of the information becomes more complicated than if there is only a single record, and there are risks of introducing inconsistencies. Likewise, as we noted earlier, it would be very inconvenient if, each time we encountered a given person, we made a new mental record of that person without any integration with previous records. The process of recognising people and other entities may be seen to be largely a process of compressing information \citep{watanabe_article_1972}.

However, it is standard practice to keep backup copies of databases and other computer information to guard against catastrophic loss of that information. And multiple copies of databases may speed up processing (e.g., mirror copies of web sites). Likewise, it seems very unlikely that evolution would have endowed us with a cognitive system that allows us to keep only one mental copy of our hard-won knowledge. It seems very likely that we will have multiple copies of our knowledge to protect us against the risk of losing it. And those copies may also have a role in speeding up processing.

In general, the ICMAUS proposals are entirely compatible with the idea that computers and brains may keep multiple (redundant) copies of stored knowledge.

\subsection{Representation of knowledge}\label{representation_of_knowledge}

Within the ICMAUS framework, all kinds of information and knowledge are to be expressed as arrays or {\em patterns} of atomic {\em symbols} in one or more dimensions.\footnote{In work to date, the focus has been on one-dimensional patterns but it is envisaged that, at some stage, the concepts will be generalised to patterns in two or more dimensions. A one-dimensional array of symbols can, of course, be described as a `sequence' or `string'. However, the term {\em pattern} will normally be favoured as a reminder that the framework is intended eventually to accommodate higher-dimensional arrays.} The system may also store alignments amongst those patterns.

The main motivation for adopting this very simple, uniform format for knowledge (echoing the adoption of production systems in Soar) has been to facilitate the manipulation and integration of diverse kinds of information within a uniform processing environment. It is evident that all kinds of `raw' information (speech, music, pictures, diagrams etc) can be stored as arrays of atomic symbols (e.g., `0' and `1') in one or more dimensions. An important aim of the research has been to see whether or how, within a single framework, this very simple format would support other kinds of representation such as grammars, trees, class hierarchies and so on, without the need for specialised processing for different kinds of representation. A selection of examples is shown below.

Another motivation for adopting a simple, uniform format for all kinds of knowledge, with uniform mechanisms for processing knowledge, is that such a scheme is more likely to be compatible with mechanisms in neural tissue than a more heterogeneous system.\footnote{Although there is clearly diversity in the organisation and functioning of different parts of the brain and nervous system, it seems likely that there is also an underlying uniformity about the way information is organised and processed in neural tissue. It is this hypothesised underlying uniformity that is the focus of interest in this research.} This is considered briefly in Section \ref{symbolic_and_connectionist}, below.

Within each pattern, each symbol is merely a `mark' that can be matched in a yes/no manner with other symbols---it has no `hidden' meaning. Thus, for example, arithmetic symbols like `+' or `$\times$' may be used in ICMAUS patterns but they would not have their normal meanings (`add' and `multiply'). Any meaning that may attach to a given symbol is to be derived from its context of other symbols and not via some extrinsic process of interpretation.\footnote{The principle that symbols should have no hidden meanings has been slightly bent in current models. In the SP61 model, a distinction has been made, within each pattern, between `identification' symbols that identify the pattern and `contents' symbols that represent the substance of the pattern. This distinction may be regarded as part of the mechanism by which the system manages its knowledge. As such, it is rather different from meanings associated with symbols like `+' or `$\times$' which are part of the knowledge itself.}

The `granularity' of symbols in the ICMAUS framework is undefined. Symbols may be used to represent very fine-grained details of a body of information (e.g., binary digits) or they may be used to represent relatively large chunks of information.

Although concepts from mathematics, logic or related disciplines such as theoretical linguistics cannot be used directly in ICMAUS representations, it is anticipated that such constructs may be modelled within the ICMAUS framework (as described in Section \ref{computing_maths_logic}, below). Even such seemingly basic constructs as `variable' or `negation' are excluded from the ICMAUS framework. But it is possible to model these constructs within the framework, as we shall see.

\subsection{Multiple alignment}\label{multiple_alignment}

In bio-informatics, a `multiple alignment' is an arrangement of two or more symbol sequences, one above the other, so that, by judicious stretching of sequences where appropriate, symbols that match each other from one sequence to another are arranged in vertical columns. A `good' alignment is one with a relatively large number of matching symbols. This kind of analysis, applied to sequences of DNA bases or amino acid residues, can be helpful in elucidating the structure, function or evolution of the corresponding molecules.

In the ICMAUS scheme, this idea has been adapted in the following ways:

\begin{itemize}

\item One of the patterns is designated `New' and the others are `Old'.

\item A `good' alignment is one which allows the New pattern to be encoded economically in terms of the Old patterns, as will be explained.

\item By contrast with multiple alignments in bio-informatics, any one pattern may appear two or more times in one alignment. As will be explained (Section \ref{pcs_for_unary_numbers}), this is not the same as allowing two or more copies of a pattern to appear within an alignment.

\end{itemize}

These points are illustrated and further explained in the examples presented throughout the article.

\subsection{The SP61 computer model}\label{SP61_model}

\sloppy The ICMAUS framework is partially realised in the SP61 computer model, described most fully in \citet{wolff_2000}. Source code for the model has now been released under the terms of the GNU General Public License and may be obtained (with the executable code) from www.informatics.bangor.ac.uk/$\sim$gerry/sp\_summary.html.

The model does not attempt any learning: it is designed to process a New pattern and to compress it, as far as possible, in terms of previously-stored Old patterns, forming multiple alignments in the course of this processing. All the alignments shown in this article are output from the SP61 model.

With regard to the application of MLE principles in the SP61 model, the size of $G$ does not vary because the model is not attempting any learning. This means that the process of seeking to minimise $(G + E)$ can be reduced to a process of seeking to minimise $E$.

At the core of the SP61 model is a process for finding full matches between patterns or good partial matches \citet{wolff_1994_scaleable}. The process is a refined version of dynamic programming \citep[see, for example,][]{sankoff_kruskall_1983} with advantages compared with standard methods: it can process patterns of arbitrary length without excessive demands for memory, it can find two or more alternative alignments for any given set of patterns, and the thoroughness of searching can be controlled by parameters.

Within the SP61 model, the matching process is applied repeatedly so that the system can build up the two or more `levels' of each multiple alignment in a pairwise manner. An outline of the model is presented in Figure \ref{SP61_figure}. In this description, `unification' of an alignment means collapsing each column of matching symbols into a single symbol so that the whole alignment can be treated as a simple sequence of symbols.

\begin{figure}[!bhpt]
\begin{center}
\begin{BVerbatim}
SP61()
{
     1 Read one or more patterns into Old and
          one pattern into New.
     2 Assign the pattern in New as the first
          and only member of a list of 'driving'
          patterns.
     3 Identify the patterns in Old as the
          initial list of 'target' patterns.
          As processing proceeds, the list of
          target patterns will be augmented with
          newly-formed alignments that have been
          unified so that each one can be treated
          as a simple sequence of symbols.
     4 while (new alignments are being formed) 
          COMPRESS() 
}
\end{BVerbatim}
\end{center}
\caption{The organisation of the SP61 model.}
\label{SP61_figure}
\end{figure}

\begin{figure}[!bhpt]
\begin{center}
\begin{BVerbatim}
COMPRESS()
{
     1 while (the end of the current list of
          driving patterns has not been reached)
     {
          1.1 Select the first or next driving
               pattern from the current list.
          1.2 Search for one or more 'good'
               alignments, each one between the
               selected driving pattern and any
               one of the current set of target
               patterns. The way in which
               alignments are evaluated in terms
               of compression is described in
               the text.
          1.3 Unify each of the new alignments so
               that it can be treated as a single
               sequence of symbols. The structure
               of the alignment is preserved to
               facilitate printing.
          1.4 Each newly unified alignment is kept
               in a temporary store until the end
               of this iteration.
     }

     2 Add the newly-formed unified alignments to a
          cumulative list of alignments formed by
          the program. This list of alignments forms
          part of the list of target patterns,
          together with the original patterns in Old.
     3 Empty the list of driving patterns and then
          select a few of the best of the newly-formed 
          unified alignments and add them to the
          list of driving patterns, ready for the next
          iteration of COMPRESS(). Although these
          patterns are identified as driving patterns
          they also retain their status as target
          patterns.
}
\end{BVerbatim}
\end{center}
\caption{The organisation of the {\em compress()} function in the SP61 model (Figure \ref{SP61_figure}).}
\label{compress_figure}
\end{figure}

In operation 1.2 in Figure \ref{compress_figure} each alignment is evaluated in terms of IC. From each alignment, a `code' can be derived as explained in Section \ref{encoding_of_new}, below. This code is a compressed representation of the New pattern in the alignment. The `compression score' for an alignment is $N_r - N_e$, where $N_r$ is the number of bits needed to express the New pattern in its `raw', unencoded state, and $N_e$ is the number of bits needed to express the encoded form of the New pattern.

Adjustments are also made to allow for `gaps' in alignments---sequences of one or more unmatched symbols. Alignments with many gaps or large ones have lower scores than alignments with gaps that are few and small. Details of how alignments are scored are given in \citet{wolff_2000}.

\subsubsection{Computational complexity}\label{computational_complexity}

Notwithstanding the astronomically large search spaces that are the rule for most multiple alignments, the heuristic techniques used in the SP61 model mean that its computational complexity is within acceptable polynomial limits. In a serial processing environment, the time complexity is estimated to be O$(log_2 n \times nm)$ where $n$ is the length of the pattern in New (in bits) and $m$ is the sum of the lengths of the patterns in Old (in bits). In a parallel processing environment, it is estimated that the time complexity would be O$(log_2 n \times n)$, depending on how well the parallelism is applied. The space complexity in both types of processing environment is estimated to be O$(m)$.

\subsection{The SP70 computer model}\label{sp70_model}

SP70 is a successor to the SP61 model designed to achieve unsupervised inductive learning by adding New patterns (or parts of them) to Old, creating appropriate `codes' (as described below), and sifting out `good' patterns from `bad' ones (in accordance with MLE principles). This model is able to abstract plausible grammars from appropriate data but more work is needed to realise the full potential of the model in this area.

Incidentally, the method of scoring alignments in SP70 does not require the adjustment for gaps required in SP61 (described in Section \ref{SP61_model}, above). This is because SP70 responds to gaps by creating new patterns, each with its own code symbols, and adding these patterns to Old. The newly-created code symbols can be used for scoring alignments containing gaps.

An outline of the model is given in Section \ref{features_of_sp70}. A relatively full description of the current version may be found in \citep{wolff_unsupervised_learning}.

\subsection{Information compression and probabilities}\label{compression_and_probabilities}

There is an intimate connection between IC and concepts of  frequency, probability and probabilistic inference \citep{solomonoff_1964, solomonoff_1986, solomonoff_1997, chater_1996, chater_1999}. Compression techniques like Huffman coding and Shannon-Fano-Elias coding \citep[see, for example,][]{cover_thomas_1991} are founded on concepts of probability. And estimates of absolute and relative probability may be derived from measures of compression.

Thus the ICMAUS framework, as an expression of IC, is fundamentally probabilistic. For this reason, it can support probabilistic inferences of various kinds, as outlined in Section \ref{probabilistic_reasoning}, below.

Each pattern in Old is assigned a frequency of occurrence in some domain and, from these values, SP61 can calculate an absolute probability for each alignment, relative probabilities for sets of alternative alignments and probabilities for inferences that may be drawn from alignments.\footnote{In SP70, the frequencies of patterns are derived from the learning process.} The methods of calculation are described in \citet{wolff_1999_prob}.\footnote{In brief, the absolute probability of any alignment is calculated as $p = 2^{-L}$, where $L$ is the length, in bits, of the code derived from the alignment as outlined in Section \ref{encoding_of_new}. For any given set of $n$ alternative alignments, $a_1 ... a_n$, where each alignment provides a match for a given set of symbols from New, the relative probability of the $j$th alignment may be calculated as $r_j = p_j / \sum_{1}^{n} p_i$. Probabilities of patterns and symbols within alignments may be derived from these relative probabilities.}

\subsection{Symbolic and connectionist processing}\label{symbolic_and_connectionist}

Although the framework has a symbolic flavour and is currently implemented as an ordinary software model, the framework is intended to express abstractions at a higher level than the distinction between symbolic and connectionist models.

It seems possible that the framework could be realised using connectionist mechanisms. Each pattern could perhaps be realised as a Hebbian `cell assembly' while each symbol might be realised either as a single nerve cell or as a small cell assembly. Detailed discussion of these possibilities is outside the scope of the present article. Current thinking in this area is described in \citet{wolff_icmaus_neural}.

\section{Natural Language Processing}\label{nl_processing}

Figure \ref{alignment_1} shows how the sentence `s i x o f t h e m d o' (that is presented to the system as New) can be aligned with patterns (in Old) representing grammatical structures.\footnote{The grammatical patterns in Old are intended to represent knowledge that has been accumulated by the system as a result of learning. In SP70, the system builds these kinds of patterns for itself. But because SP61 does not attempt learning, the patterns are simply given to the system at the start of processing.} By convention in this research, New is always shown in the top row of each alignment with patterns from Old in the rows underneath. The order of the rows below the top row is entirely arbitrary and has no special significance.\footnote{This example and others in this article have been deliberately chosen to be relatively simple. This is partly for the sake of clarity in exposition and partly because complex alignments can be difficult to display within the confines of the printed page. As already indicated (Section \ref{computational_complexity}), the computational complexity of the model is acceptable. The model readily produces more complex alignments where appropriate. Relatively elaborate examples of multiple alignment may be found in \citet{wolff_2000}.}

\begin{figure}[!bhpt]
\fontsize{07.00pt}{08.40pt}
\begin{center}
\begin{BVerbatim}
0                s i x          o f           t h e m                  d o       0
                 | | |          | |           | | | |                  | |      
1                | | |          | |    N Np 0 t h e m #N               | |       1
                 | | |          | |    |              |                | |      
2                | | |    Q P   | | #P N              #N #Q            | |       2
                 | | |    | |   | | |                    |             | |      
3                | | |    | P 2 o f #P                   |             | |       3
                 | | |    |                              |             | |      
4         N Np 1 s i x #N |                              |             | |       4
          | |          |  |                              |             | |      
5      NP N |          #N Q                              #Q #NP        | |       5
       |    |             |                                  |         | |      
6      |    |             |                                  |  V Vp 2 d o #V    6
       |    |             |                                  |  | |        |    
7 S    NP   |             |                                 #NP V |        #V #S 7
  |    |    |             |                                       |           | 
8 S PL NP   Np            Q                                       Vp          #S 8
\end{BVerbatim}
\end{center}
\caption{The best alignment found by SP61 with `s i x o f t h e m d o' in New and grammatical patterns in Old.}
\label{alignment_1}
\end{figure}

In the alignment, most of the patterns representing grammatical structures are similar to rules in a context-free phrase-structure grammar (CF-PSG). For example, the pattern `S NP \#NP V \#V \#S' expresses the idea that a (simple) sentence is composed of a noun phrase followed by a verb. In a CF-PSG, this would be expressed by a rule like `S $\rightarrow$ NP V'. The main differences between the two representations are that the patterns do not contain rewrite arrows and that each structure is normally referenced by an initial symbol (e.g., `NP') and a termination symbol (e.g., `\#NP').

In the pattern `S NP \#NP V \#V \#S', each pair of symbols `NP \#NP' and `V \#V' functions as a `variable' that can receive an appropriate value, as shown in the alignment in Figure \ref{alignment_1}.

\subsection{Discontinuous syntactic agreements}\label{syntactic_agreements}

If we ignore row 8, the alignment in Figure \ref{alignment_1} maps fairly directly onto the kind of parsing one would obtain with a CF-PSG. That kind of parsing expresses the division of the sentence into `chunks' of structure such as phrases and words but it does not show the kind of dependency or `agreement' that exists between the number of the subject (singular or plural) and the number of the verb. This kind of agreement is `discontinuous' in the sense that it can bridge arbitrarily large amounts of intervening structure.

This aspect of the structure of natural languages has, for several years, been handled quite well by various grammatical systems including Transformational Grammar, Definite Clause Grammars \citep{pereira_warren_1980}, FAPR \citep[described in][]{gazdar_mellish_1989} and others.

Figure \ref{alignment_1} illustrates an alternative way in which syntactic agreements may be expressed which is, as far as I can ascertain, significantly different from any existing system. The agreement between the plural subject (`s i x') and the plural verb (`d o') is expressed in row 8 of the alignment in a relatively simple and direct manner with the pattern `S PL NP Np Q Vp \#S '. The symbols `S' and `\#S' serve to tie this pattern into the main sentence pattern; `PL' at the beginning marks the sentence as plural (and has a use which is described shortly); the symbols `Np' and `Vp' express the connection between the plural noun at the beginning of the sentence and the verb at the end; and the symbol `Q' is needed to show that the `Np' in this plural pattern refers to the noun before the qualifying phrase, not the plural noun (`t h e m') within that phrase.

More elaborate examples of this kind of parsing by multiple alignment may be found in \citet{wolff_2000}, including one example that shows overlapping patterns of number dependency and gender dependency in French, and another showing the interesting pattern of interlocking dependencies in English auxiliary verbs.

\subsection{Encoding of New in terms of patterns in Old}\label{encoding_of_new}

Each of the Old patterns in Figure \ref{alignment_1} contains symbols that may be used as an identifier or `code' for the pattern. For example, the pattern `N PL 0 t h e m \#N' may be coded using the symbols `N PL 0 \#N'. These code symbols provide the means of expressing New in a relatively succinct form in terms of the Old patterns in the alignment.

Looking at the alignment, one might think that relatively large numbers of code symbols would be needed to encode the whole sentence. But many of the code symbols (e.g., `N' and `\#N') merely serve to tie patterns together within the alignment and do not need to be included in the encoding of the sentence. In an example like the one shown in Figure \ref{alignment_1}, a code for the sentence may be constructed by selecting the symbols in the alignment that are not matched to any other symbol. In this example, the code created in this way is `PL 1 2 Np 0 2'. As we shall see (next), this code can be used to reconstruct the original sentence.

\subsection{Production of language}

An interesting feature of the ICMAUS framework (and the SP61 model) is that, without any modification, it can support the production of language as well as the parsing of language. This can be done by replacing the sentence in New with an encoding of the sentence as just described. With this substitution, the best alignment found by SP61 is shown in Figure \ref{alignment_2}.

\begin{figure}[!bhpt]
\fontsize{07.00pt}{08.40pt}
\begin{center}
\begin{BVerbatim}
0   PL         1              2          Np 0                        2           0
    |          |              |          |  |                        |          
1   |          |            P 2 o f #P   |  |                        |           1
    |          |            |       |    |  |                        |          
2   |     N Np 1 s i x #N   |       |    |  |                        |           2
    |     | |          |    |       |    |  |                        |          
3   |  NP N |          #N Q |       |    |  |            #Q #NP      |           3
    |  |    |             | |       |    |  |            |   |       |          
4   |  |    |             Q P       #P N |  |         #N #Q  |       |           4
    |  |    |             |            | |  |         |      |       |          
5 S |  NP   |             |            | |  |         |     #NP V    |     #V #S 5
  | |  |    |             |            | |  |         |         |    |     |  | 
6 | |  |    |             |            N Np 0 t h e m #N        |    |     |  |  6
  | |  |    |             |                                     |    |     |  | 
7 | |  |    |             |                                     V Vp 2 d o #V |  7
  | |  |    |             |                                       |           | 
8 S PL NP   Np            Q                                       Vp          #S 8
\end{BVerbatim}
\end{center}
\caption{The best alignment found by SP61 with `PL 1 2 Np 0 2' in New and with the same grammatical patterns in Old as were used for Figure \ref{alignment_1}.}
\label{alignment_2}
\end{figure}

If this alignment is unified, and if we ignore the code symbols in the unified pattern, we obtain the words of the original sentence in their correct order. In effect, the system achieves the apparent paradox of `decompression by compression'. A relatively full discussion of this point and how the apparent paradox is resolved may be found in \citet{wolff_2000}.

Notice how the inclusion of the single symbol `PL' in the encoding of the sentence eliminates the need to include the two symbols `Np' and `Vp'. This is how patterns like the one shown in row 8 of the two alignments can contribute to IC.

\subsection{Ambiguity}

It should be emphasised that SP61 normally forms several alternative alignments for a given set of patterns, each one with its own compression score. Amongst these alternatives are many that look, intuitively, wrong. Almost always, the one with the highest score (as in Figure \ref{alignment_1}) is also the one which is, intuitively, fully correct or most nearly correct. Where there is syntactic ambiguity, SP61 will normally deliver two (or more) alignments with relatively high scores.

\subsection{Integration of syntax and semantics}

The way in which the ICMAUS framework might support the integration of syntax and semantics has not yet been examined in any detail. However, as was noted above, the very simple format for knowledge that has been adopted was chosen deliberately with the aim of facilitating the representation of diverse kinds of knowledge and their integration.

Examples presented below show how the framework can support aspects of non-syntactic knowledge such as class hierarchies with inheritance of attributes (Section \ref{class_hierarchies}) and polythetic categories (Section \ref{polythetic_classes}).

\section{Computing, Mathematics and Logic}\label{computing_maths_logic}

Up until the 1940s, `computing' was something done almost exclusively by people. In those days, it would have seemed entirely natural to find common ground between the psychological processes involved in computation and other aspects of human cognition. Indeed, a consideration of the elements of human computation provided part of the inspiration for Alan Turing's brilliantly-conceived `Universal Turing Machine'. And of course George Boole regarded mathematical theories of logic and probability as being founded on ``Laws of Thought''. The ICMAUS proposals may be seen as an extension of these lines of thinking.

In general, the ICMAUS framework is probabilistic (Section \ref{compression_and_probabilities}). Given appropriate input, it is capable of finding good partial matches between patterns (illustrated in Section \ref{pattern_recognition}, below) and it is typically able to find two or more alternative alignments for any given set of patterns in New and Old. But constraints may be applied or the focus may be narrowed to consider only those alignments in which each symbol in New has been matched to at least one symbol in Old or to consider only the `best' alignment for any given set of patterns. With these kinds of constraint, the system may be used to model `exact' forms of calculation and reasoning.

\subsection{Modelling a Universal Turing Machine and Post Canonical System}

For about 60 years, the concept of a `Universal Turing Machine' (UTM), and equivalent models such as the `Post Canonical System' (PCS), have provided an abstract definition of the concept of `computing'. These models have been very successful but have little to say about things like learning, fuzzy pattern recognition, probabilistic reasoning and other topics in cognitive science and artificial intelligence. A motivation for developing the ICMAUS framework is to see whether or how it may be possible to plug this gap.

In \citet{wolff_1999_prob}, I have argued that the operation of a PCS may be interpreted within the ICMAUS framework. Since any UTM can be modelled by a PCS \citep[see][]{minsky_1967}, we may also argue that the UTM can be modelled within the ICMAUS framework.

\subsubsection{Example: a PCS for the generation of unary numbers}\label{pcs_for_unary_numbers}

A PCS comprises an alphabet of symbols, one or more `primitive assertions' or `axioms', and one or more `productions'. Figure \ref{PCS_unary_figure} shows a PCS for the generation of unary numbers (where 0 = 0, 1 = 01, 2 = 011, 3 = 0111, and so on). Such a system can be run `forwards' to create unary numbers or `backwards' to recognise unary numbers.

\begin{figure}[!bhpt]
\begin{quotation}
\noindent {\em Alphabet}: the symbols 0 and 1.

\noindent {\em Axiom}: 0.

\noindent {\em Production}: If any string `\$' is a number, then so is the string `\$ 1'. \newline \indent This can be expressed with the rule: \$ $\rightarrow$ \$ 1.
\end{quotation}
\caption{A PCS to generate unary numbers.}
\label{PCS_unary_figure}
\end{figure}

This example can be modelled in the ICMAUS framework if we represent the axiom with the pattern `X a 0 \#X' and the production with the pattern `X b X \#X 1 \#X'. The symbols `X' and `\#X' at the beginning and end of each pattern are code symbols needed to tie the patterns together in an alignment. The symbols `a' and `b' are code symbols needed for the scoring system in SP61. In the second of these two patterns, the pair of contiguous symbols `X \#X' may be read as ``any number'' and the whole pattern may be read as ``a number comprises any number followed by `1'''. As in our example of parsing by multiple alignment (Section \ref{nl_processing}), a pair of contiguous symbols like `X \#X' can imitate the effect of a variable.

\begin{figure}[!bhpt]
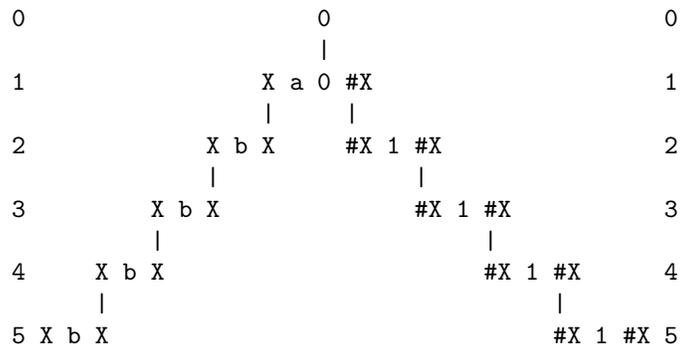

\begin{center}
\begin{BVerbatim}
0                     0    1    1    1    1    0
                      |    |    |    |    |   
1                 X a 0 #X |    |    |    |    1
                  |     |  |    |    |    |   
2             X b X     #X 1 #X |    |    |    2
              |              |  |    |    |   
3         X b X              #X 1 #X |    |    3
          |                       |  |    |   
4     X b X                       #X 1 #X |    4
      |                                |  |   
5 X b X                                #X 1 #X 5

(a)

0                     0                        0
                      |                       
1                 X a 0 #X                     1
                  |     |                     
2             X b X     #X 1 #X                2
              |              |                
3         X b X              #X 1 #X           3
          |                       |           
4     X b X                       #X 1 #X      4
      |                                |      
5 X b X                                #X 1 #X 5

(b)
\end{BVerbatim}
\end{center}
\caption{(a) Recognition of a unary number as a multiple alignment. (b) Production of a unary number in the same terms.}
\label{alignment_3}
\end{figure}

Figure \ref{alignment_3} (a) shows the best alignment found by SP61 with the unary number `0 1 1 1 1' in New and patterns for the axiom and the production in Old. The alignment confirms that the number conforms to the rules of unary arithmetic and thus, in effect, recognises the number as a unary number. Figure \ref{alignment_3} (b) shows one of many alternative alignments produced by SP61 with `0' in New and the same patterns in Old. Ignoring the `service' symbols, the alignment contains the same symbols as before, in the same order. In effect, the system has produced the unary number `0 1 1 1 1'.

Notice that, in each of the two alignments shown in Figure \ref{alignment_3}, the pattern `X b X \#X 1 \#X' appears several times. As was indicated in Section \ref{multiple_alignment}, there is an important difference between two or more appearances of a pattern in one alignment and two or more copies of a pattern in one alignment. In the first case, any one symbol in one appearance should not be matched with the corresponding symbol in another appearance-because this means matching the given symbol with itself. In the latter case, no such restrictions apply.\footnote{A New pattern may be a copy of an Old pattern but, in general, Old should not contain multiple copies of any pattern.}

\subsubsection{Simplicity and power}

The main difference between the ICMAUS framework and earlier models is in the provision of a capability for finding good partial matches between patterns and the ability to build multiple alignments.

Although the ICMAUS framework is not as simple as the UTM or PCS models, it seems that the added complexity is more than offset by an increase in descriptive or explanatory power, particularly in areas of interest within cognitive science and artificial intelligence.

\subsection{Mathematics and logic}

This subsection presents three examples to illustrate the way in which the ICMAUS framework may be used to model concepts in mathematics and logic. Much fuller discussion and more examples may be found in \citet{wolff_maths_logic}.

\subsubsection{Adding two numbers}

The four patterns shown in Figure \ref{one_bit_adder} define a function that, in computer science jargon, is known as a `half adder'. In each row of the figure, the two digits between `A' and `S' are two one-bit numbers, the digit following the `S' is the sum of those two numbers and the digit following the `C' is the `carry out' bit.

\begin{figure}
\begin{center}
\begin{tabular}{l}
A 1 1 S 0 C 1 \\
A 1 0 S 1 C 0 \\
A 0 1 S 1 C 0 \\
A 0 0 S 0 C 0 \\
\end{tabular}
\end{center}
\caption{A set of patterns defining a half adder.}
\label{one_bit_adder}
\end{figure}

Figure \ref{alignment_4} shows the best alignment found by SP61 with `A 0 1 S C' in New and the patterns from Figure \ref{one_bit_adder} in Old. In effect, the pattern in New is a request to add the digits `0' and `1'. The result of the computation is the two unmatched digits in the alignment, `1' for the sum and `0' for the carry out bit. This is, of course, the correct result for the addition of `0' and `1'.

\begin{figure}
\begin{center}
\begin{BVerbatim}
0 A 0 1 S   C   0
  | | | |   |  
1 A 0 1 S 1 C 0 1
\end{BVerbatim}
\end{center}
\caption{The best alignment found by SP61 with `A 0 1 S  C' in New and the patterns shown in Figure \ref{one_bit_adder} in Old. The digits at the ends of the rows are row numbers, not part of the figure itself.}
\label{alignment_4}
\end{figure}

In order to progress beyond the addition of two one-bit numbers, it is necessary to be able to combine functions so that the output of one function becomes the input of another. In order to add numbers containing two or more bits, the set of patterns shown in Figure \ref{one_bit_adder} needs to be augmented so that it becomes a `full adder' in which there is a `carry in' bit as well as a carry out bit, and the  ICMAUS framework needs to be applied recursively. This is explained, with an example from SP61, in \citet{wolff_maths_logic}.

The way in which the output of one function can become the input of another is shown next in the domain of logic.

\subsubsection{Combining logical functions}

The first four patterns in Figure \ref{notxor_figure} define the XOR logical function: the first two digits in each pattern are the digits to be processed and the digit between `A' and `\#A' is the result.\footnote{The function is the same as the half adder but without the carry out bit.}

\begin{figure}
\begin{center}
\begin{tabular}{l}
XOR 1 1 A 0 \#A \\
XOR 1 0 A 1 \#A \\
XOR 0 1 A 1 \#A \\
XOR 0 0 A 0 \#A \\
A 1 \#A NOT R 0 \#R \\
A 0 \#A NOT R 1 \#R \\
NOTXOR XOR NOT R \#R \#NX \\
\end{tabular}
\end{center}
\caption{Patterns for the XOR function, the NOT function and the NOTXOR function.}
\label{notxor_figure}
\end{figure}

The next two patterns in the figure define the NOT logical function: the first digit in each pattern is the input and the last digit is the output. The last pattern in the figure serves to combine the two functions so that the output of the XOR function can become the input to the NOT function.

Figure \ref{alignment_5} shows the best alignment found by SP61 with the pattern `NOTXOR 0 1 A \#NX' in New and the patterns from Figure \ref{notxor_figure} in Old. As can be seen in the alignment, the result of the XOR computation (`1') appears between the `A' and `\#A' columns in the middle of the alignment and this becomes the input to the NOT function. The overall result (`0') is shown between the `R' and `\#R' columns in the alignment.

\begin{figure}
\begin{center}
\begin{BVerbatim}
0 NOTXOR     0 1 A                 #NX 0
    |        | | |                  | 
1   |    XOR 0 1 A 1 #A             |  1
    |     |      | | |              | 
2   |     |      A 1 #A NOT R 0 #R  |  2
    |     |              |  |   |   | 
3 NOTXOR XOR            NOT R   #R #NX 3
\end{BVerbatim}
\end{center}
\caption{The best alignment found by SP61 with `NOTXOR 0 1 A \#NX' in New and the patterns from Figure \ref{notxor_figure} in Old.}
\label{alignment_5}
\end{figure}

This example is not as general as conventional functions (where there is no need to match the name of the output area of one function with the input area of another) but it does, nevertheless, demonstrate how, within the ICMAUS framework, information can be passed from one function to another. It is anticipated that, when SP70 is more fully developed, it will allow greater generality in this area.

\subsubsection{Syllogistic reasoning}

Consider the following text-book example of a {\em modus ponens} syllogism:

\begin{enumerate}

\item All humans are mortal.

\item Socrates is human.

\item Therefore, Socrates is mortal.

\end{enumerate}

In logical notation, this may be expressed as:

\begin{enumerate}

\item $\forall x: human(x) \Rightarrow mortal(x)$.

\item $human(Socrates)$.

\item $\therefore mortal(Socrates)$.

\end{enumerate}

In the ICMAUS framework, the first of these propositions may be expressed with the pattern `X \#X human true $\Rightarrow$ mortal true'. In the manner of Skolemization, the variable `X \#X' in the pattern represents anything at all and may thus be seen to be universally quantified. The scope of the variable may be seen to embrace the entire pattern, without the need for it to be repeated. In keeping with the earlier remarks about ``no hidden meanings'' (Section \ref{representation_of_knowledge}), the symbol `$\Rightarrow$' in the ICMAUS pattern serves simply as a separator between `human true' and `mortal true'.

If this pattern is included (with other patterns) in Old and if New contains the pattern `Socrates human true $\Rightarrow$' (corresponding to `human(Socrates)' and, in effect, a request to discover what that proposition implies), SP61 finds one alignment (shown in Figure \ref{alignment_6}) that encodes all the symbols in New. After unification, this alignment may be read as a statement that because it is true that Socrates is human it is also true that Socrates is mortal.

\begin{figure}
\begin{center}
\begin{BVerbatim}
0   Socrates    human true =>             0
       |          |    |   |             
1 X    |     #X human true => mortal true 1
  |    |     |                           
2 X Socrates #X                           2
\end{BVerbatim}
\end{center}
\caption{The best alignment found by SP61 with `Socrates human true $\Rightarrow$' in New and patterns in Old that include `X \#X human true $\Rightarrow$ mortal true'.}
\label{alignment_6}
\end{figure}

\section{Pattern Recognition and Best-Match Information Retrieval}\label{pattern_recognition}

The dynamic programming built into the SP61 model means that it can find good partial matches between a New pattern and one or more stored patterns. It can thus mimic the kind of `fuzzy' pattern recognition and best-match information retrieval that is such a prominent feature of human perception and cognition (Section \ref{fuzzy_matching}, next). The ability to form multiple alignments allows the modelling of class hierarchies with inheritance of attributes (Section \ref{class_hierarchies}) and the modelling of polythetic categories (Section \ref{polythetic_classes}).

\subsection{Fuzzy matching}\label{fuzzy_matching}

Figure \ref{alignment_7} shows how the system can achieve fuzzy recognition of words. The figure shows, in descending order of their compression scores, the best five alignments found by SP61 with the misspelled word `i m f o r m t i x o n' in New and a selection of correctly-spelled words in Old. Notice how the system can accommodate errors of omission, commission and substitution.\footnote{Given that the last four alignments shown in Figure \ref{alignment_7} each have 6 matching pairs of symbols, one might expect them all to have the same compression score. The reason their scores are different is that the compression score takes account of unmatched symbols (`gaps') in the alignments. Alignments with many gaps or large gaps have lower scores than those with gaps that are few and small.}

\begin{figure}
\begin{center}
\begin{BVerbatim}
0     i m f o r m   t i x o n     0
      |   | | | |   | |   | |    
1 %I3 i n f o r m a t i   o n #I3 1

0 i m f o r m   t i x o n   0
      | | | |   | |        
1 %F1 f o r m a t i v e #F1 1

0 i m f o r m         t i x o n     0
        |   |         | |   | |    
1 %C1 c o   m p u t a t i   o n #C1 1

0 i m f o r m     t     i x o n     0
        |   |     |     |   | |    
1 %C1 c o   m p u t a t i   o n #C1 1

0     i         m f o r m t i x o n     0
      |         |         | |   | |    
1 %I3 i n f o r m a       t i   o n #I3 1
\end{BVerbatim}
\end{center}
\caption{The five best alignments found by SP61 with `i m f o r m t i x o n' in New and a selection of correctly spelled words in Old. The alignments are shown in descending order of their compression scores.}
\label{alignment_7}
\end{figure}

Of course, this kind of pattern recognition is done quite well by systems for spelling checking and correction. What is different about the ICMAUS framework is the range of its capabilities (illustrated in this article), well beyond what can be achieved with any spelling checker. Another difference-not illustrated in this example-is that the SP61 model can bridge arbitrarily large gaps between matching sections, unlike any ordinary spelling checker.

If the kinds of capabilities already demonstrated can be generalised to the processing of patterns in two dimensions, they should prove useful in such tasks as visual pattern recognition and scene analysis where it often happens that patterns and objects are partially obscured, one behind another.

\subsection{Class hierarchies and inheritance of attributes}\label{class_hierarchies}

Apart from allowing fuzzy recognition or retrieval of patterns, the ICMAUS framework allows recognition to occur through two or more levels of abstraction. Figure \ref{alignment_8} shows a simple example where symbols for `sustains\_life', `long' and `crusty' serve to identify an object as being a `baguette' which, as such, is a form of `bread', which is a form of `food'.\footnote{Compared with the previous alignments in this article, the alignment shown in Figure \ref{alignment_8} is rotated by $90^o$, with New in the left-most column and patterns from Old in the columns to the right. This arrangement allows the alignment to be fitted more easily on to the printed page.} As a baguette, the object is `white' as well as being `long' and `crusty'. As bread, it is made of `flour', `yeast' and `water'. And as food, it is `organic' (derived from living things) and contains `fat', `protein' and `carbohydrate'. The proportions of these substances (`small', `medium', `large') are recorded at the level of bread.

\begin{figure}
\fontsize{09.00pt}{10.80pt}
\begin{center}
\begin{BVerbatim}
                baguette                                 
                bread ----- bread                        
                            food ---------- food         
                            fat ----------- fat          
                            small                        
                            #fat ---------- #fat         
                            protein ------- protein      
                            medium                       
                            #protein ------ #protein     
                            carbohydrate -- carbohydrate 
                            large                        
                            #carbohydrate - #carbohydrate
sustains_life ----------------------------- sustains_life
                                            organic      
                            #food --------- #food        
                            flour                        
                            yeast                        
                            water                        
                #bread ---- #bread                       
long ---------- long                                     
                white                                    
crusty -------- crusty                                   
                #baguette                                
\end{BVerbatim}
\end{center}
\caption{The best alignment formed by SP61 with symbols representing attributes of food in New and patterns for classes of food in Old. In this arrangement, New is in the left-most column, with patterns from Old in the columns to the right.}
\label{alignment_8}
\end{figure}

In the jargon of object-oriented design, the patterns for `food', `bread' and `baguette' represent `classes' of entity that are arranged in a `hierarchy` (with 'food' at the top in this case and `baguette' at the bottom). And `attributes' at each level (e.g., `white', `made of flour', `contains protein' etc) are `inherited' by all the lower levels.

\subsubsection{Class hierarchies and inheritance of attributes in human cognition}

\citet{collins_quillian_1969} suggested that these principles operate in human perception and cognition, and their experimental results seemed to support the proposal. However, later experimental work cast doubt on the validity of those early results \citep[see, for example,][p. 178 ff.]{barsalou_1992}.

A full discussion of this issue is not possible here. Class hierarchies with inheritance of attributes are such powerful ideas that it seems unlikely that our brains would not exploit them. It is hard to believe, that, for every person we know, we keep a complete record of all their human attributes. It seems much more likely that knowing that someone is a person allows us to infer most of their attributes from our general knowledge of people.

Bearing in mind the high levels of parallel processing that operate in the brain, and bearing in mind the probability that our mental knowledge is compressed and redundant at the same time (Section \ref{compressed_and_redundant}), it is very difficult to draw sound conclusions from reaction times and similar experimental measurements.

\subsection{Polythetic classes}\label{polythetic_classes}

People not only have an ability to recognise things despite errors of omission, commission and substitution (as in the example in Section \ref{fuzzy_matching}) but it seems that many of our concepts express a `family resemblance' or are `polythetic', meaning that no single attribute need appear in all members of the class. Although, in a na\"{i}f view of the nature of concepts, this aspect of `natural' categories may be puzzling, it can actually be modelled very simply by the use of re-write rules. For example, the following five rules: `1 $\rightarrow$ 2 3'; `2 $\rightarrow$ A'; `2 $\rightarrow$ B'; `3 $\rightarrow$ C'; and `3 $\rightarrow$ D', define the polythetic class {`AC', `AD', `BC', `BD'}. Notice that none of the attributes `A', `B', `C' or `D' appear in all members of the class.

Given that these kinds of re-write rules can be modelled in the ICMAUS framework (as we saw in Section \ref{nl_processing}), it is clear that the framework can accommodate this aspect of human thinking.

\section{Probabilistic reasoning}\label{probabilistic_reasoning}

There is now a vast literature relating to probabilistic reasoning: standard parametric and non-parametric statistics; {\em ad hoc} uncertainty measures in early expert systems; Bayesian statistics; Bayesian/belief/causal networks; Markov networks; Self-Organising Feature Maps; fuzzy set theory and `soft computing'; the Dempster-Shaffer theory; abductive reasoning; reasoning with default values and nonmonotonic reasoning; autoepistemic logic, defeasible logic, probabilistic, possibilistic and other kinds of logic designed to accommodate uncertainty; MLE; algorithmic probability and algorithmic complexity theory; truth maintenance systems; decision analysis; utility theory; and so on.

No attempt can be made here to compare the ICMAUS proposals with these many alternatives. This section merely presents some examples suggesting how the framework can support probabilistic reasoning.

The example shown in Figure \ref{alignment_8} illustrates the way in which inferences may be drawn from the kinds of multiple alignment formed by SP61. As in the examples shown in Figure \ref{alignment_4}, Figure \ref{alignment_5} and Figure \ref{alignment_6}, the columns of any alignment that are not aligned with any of the symbols in New may be seen to be inferences that may be drawn from the alignment. Thus, in Figure \ref{alignment_8}, we can infer that the object that has been recognised contains fat, protein and carbohydrate, that it is made of flour, yeast and water, and so on.

\subsection{Probabilistic `deduction' and abduction}

Inheritance of attributes, as just described, may be seen as a kind of `deduction': knowing the class or classes to which an object belongs allows us to deduce attributes that have not been directly observed. Quote marks are used because this kind of inference lacks the formal properties of classical kinds of deduction. And these inferences are probabilistic because they do not need to be constrained in the kinds of ways indicated in Section \ref{compression_and_probabilities}.

Figure \ref{alignment_9} shows a simple example of this kind of deduction. New (in the left column) records the fact that `Tibbs' is a `mammal' and the pattern in the third column describes the attributes of mammals. The alignment allows us to infer that ``If Tibbs is a mammal, it is likely that he or she is warm blooded, furry and so on''.

\begin{figure}
\begin{center}
\begin{BVerbatim}
mammal --------- mammal      
         name -- name        
Tibbs -- Tibbs               
         #name - #name       
                 warm_blooded
                 furry       
                 ...         
                 #mammal     
\end{BVerbatim}
\end{center}
\caption{The best alignment found by SP61 with `mammal Tibbs' in New and patterns for classes of animal in Old.}
\label{alignment_9}
\end{figure}

An attractive feature of the ICMAUS framework is that it is just as easy to make `backwards' abductive inferences as the kind of `forwards' inference just shown. Figure \ref{alignment_10} shows the two best alignments found by SP61 with `Tibbs warm\_blooded' in New and the same patterns in Old as were used for Figure \ref{alignment_9}. From these alignments, we may conclude that, as a warm blooded creature, Tibbs might be either a bird or a mammal.

\begin{figure}
\begin{center}
\begin{BVerbatim}
                       bird        
               name -- name        
Tibbs -------- Tibbs               
               #name - #name       
warm_blooded --------- warm_blooded
                       wings       
                       feathers    
                       ...         
                       #bird       

(a)

                       mammal      
               name -- name        
Tibbs -------- Tibbs               
               #name - #name       
warm_blooded --------- warm_blooded
                       furry       
                       ...         
                       #mammal     

(b)
\end{BVerbatim}
\end{center}
\caption{The two best alignments found by SP61 with `Tibbs warm\_blooded' in New and the same patterns in Old as were used for Figure \ref{alignment_9}.}
\label{alignment_10}
\end{figure}

In the example shown in Figure \ref{alignment_9}, there is only one alignment that matches all the symbols in New. So the relative probability of the alignment (and inferences that may be drawn from the alignment) is 1.0. In the second example, where there are two alignments that match all the symbols in New, the relative probabilities calculated by SP61 are 0.59 (for Figure \ref{alignment_10} (a)) and 0.41 (for Figure \ref{alignment_10} (b)). Notice that these two values sum to 1.0 because, in the `closed world' of information supplied to the system, birds and mammals are the only kinds of warm-blooded animal.\footnote{In this case, it seems also to be true of the `real' world that birds and mammals are the only warm-blooded kinds of creature. But, like any cognitive system (natural or artificial), there is no guarantee that stored knowledge will be an accurate reflection of the real world.}

\subsection{Chains of inference and other compound inferences}

Figure \ref{alignment_8} shows how the system can support inferences through two or more levels. Similar `chains' of inference can be modelled if Old contains patterns representing `if ... then' associations like `smoke fire', `night day', `black\_clouds rain' etc. 

A very simple whodunit example is shown in Figure \ref{alignment_11} (a) where New (in the left column) records an association between a `suspect' and a particular `motive', whilst Old contains, {\em inter alia}, a pattern (in the second column) recording an association between the `motive' and the `crime', a pattern (in the third column) recording the fact that the crime was committed at a particular place (`scene1') and at a particular time (`time1'), and a pattern (in the fourth column) recording the fact that the suspect was in the same place at the same time. The entire alignment may be seen as a chain of inference suggesting that the suspect may have committed the crime.

Figure \ref{alignment_11} (b) shows a similar example of `means-ends analysis': finding a (not very good) railway route from London to Edinburgh.

\begin{figure}
\fontsize{09.00pt}{10.80pt}
\begin{center}
\begin{BVerbatim}
suspect ------------------- suspect
                   scene1 - scene1 
                   time1 -- time1  
          crime -- crime           
motive -- motive                   

(a)

London ---- London                                   
            Birmingham - Birmingham                          
                         Manchester ------------- Manchester
                                      York ------ York   
Edinburgh --------------------------- Edinburgh            

(b)
\end{BVerbatim}
\end{center}
\caption{Multiple alignments representing chains of inference: (a) A whodunnit chain of inference. (b) Finding a route between London and Edinburgh.}
\label{alignment_11}
\end{figure}

Readers will notice that, in these examples and others in this article, the order of symbols in each pattern is critical. If, in the second column of Figure \ref{alignment_11} (a), the order of `crime' and `motive' had been reversed, then the system would not have been able to complete the alignment. The links in Figure \ref{alignment_11} (b) are directional: the pattern 'Manchester York' means there is a train from Manchester to York but leaves unspecified whether or not there is a train in the other direction.

Bi-directional links could, of course, be supplied explicitly. However, the redundancy in that solution can be avoided if we adopt an arbitrary order for symbols in all the patterns (e.g., alpha-numeric) and avoid clashes between symbols by tying them into a `framework' pattern like the one shown in the third column of Figure \ref{alignment_12}. Each symbol (other than the symbols used in the framework) has its own slot in the framework so that it does not interfere with other symbols.\footnote{It should be clear that bi-directional railway journeys could be accommodated in a similar way.}

\begin{figure}
\fontsize{09.00pt}{10.80pt}
\begin{center}
\begin{BVerbatim}
                      1 ---------- 1                          
accused -------------------------- accused                    
                      2 ---------- 2                          
                      3                                       
           4 -------- 4 ------------------------------ 4      
           destroy ----------------------------------- destroy
           5 -------- 5 -- 5 --------------- 5 ------- 5      
                           fire ------------ fire ---- fire   
                      6 -- 6 --------------- 6 ------- 6      
matches ------------------------------------ matches          
                      7 ---------- 7 ------- 7                
                                   petrol -- petrol           
                      8 ---------- 8 ------- 8                
                      9 -- 9                                  
smoke -------------------- smoke                              
                      10 - 10                                 
           11 ------- 11                                      
the_barn - the_barn                                           
           12 ------- 12                                      
\end{BVerbatim}
\end{center}
\caption{The best alignment found by SP61 with facts to be explained in New (on the left) and patterns for relevant facts and associations in Old.}
\label{alignment_12}
\end{figure}

In this second whodunit scenario, New records the fact that the `accused' was seen with `matches' near `the\_barn' and that `smoke' was seen at about the same time. The patterns from Old that have been tied into the alignment include ones showing:

\begin{itemize}

\item The fact that the barn was destroyed (column 2).

\item The well-known association between smoke and fire (column 4).

\item That the accused had petrol (column 5).

\item The association between fire, matches and petrol (column 6).

\item The fact that fire destroys things (column 7).

\end{itemize}

It should be clear from the examples in this section that, in general, the ICMAUS framework can accommodate arbitrary networks and trees represented by sets of patterns. It may thus be used in other applications where these kinds of representation are appropriate.

\subsection{Nonmonotonic reasoning and default values}

It has been recognised for some time that, by contrast with classical logic, everyday human reasoning allows us to modify conclusions in the light of new information. We may, for example, infer that Tweety can fly because Tweety is a bird but revise this conclusion if we subsequently learn that Tweety is a penguin \citep[see, for example,][]{ginsberg_1994, antoniou_1997}. The ICMAUS framework provides a means of modelling this kind of `nonmonotonic' or `defeasible' style of reasoning.\footnote{Owing to over-exposure in the literature, Tweety is suffering from nervous exhaustion and will, accordingly, be replaced in the following example by a loaf of bread!}

In the example illustrated in Figure \ref{alignment_8}, the colour of a baguette was given unambiguously as `white'. Instead of specifying the colour directly in this way, we can replace the symbol `white' with a pair of symbols `colour \#colour' (functioning as a variable for the colour for the loaf, unspecified as yet) and add a new pattern to Old, `standard baguette colour white \#colour \#baguette \#standard', that says, in effect, that the default value for the colour of a standard baguette is `white'. In addition, we may add the pattern `special baguette colour brown \#colour \#baguette rustic \#special' to Old that says, in effect, that there is a special `rustic' subclass of baguette where the colour is `brown'.

Figure \ref{alignment_13} shows the three best alignments formed by SP61 with the same symbols in New as before and with modifications to Old as described (and, for the sake of simplicity, without the details of fat, protein and carbohydrate).

From the first of these alignments, we may infer merely that the unknown object is a baguette and that its colour is unspecified. From the second alignment (that contains the first alignment), we may infer that the colour of the baguette could be `white'. And from the third alignment (which also contains the first alignment), we may infer that that the colour could be `brown'.

For these alignments, SP61 calculates probabilities of the inferences that may be drawn from the alignments. In this example, the calculated relative probability that the unknown object is a `baguette' is 1.0, and the values for `white' and `brown' are 0.55 and 0.45, respectively.

\begin{figure}
\fontsize{06.50pt}{07.80pt}
\begin{center}
\begin{BVerbatim}
                baguette                          
                bread ----- bread                 
                            food --- food         
                            ... ---- ...          
sustains_life ---------------------- sustains_life
                                     organic      
                            #food -- #food        
                            flour                 
                            yeast                 
                            water                 
                #bread ---- #bread                
long ---------- long                              
                colour                             
                #colour                            
crusty -------- crusty                            
                #baguette                         

(a)

                            standard                          
                baguette -- baguette                          
                bread --------------------------------- bread 
                                        food ---------- food  
                                        ... ----------- ...   
sustains_life ------------------------- sustains_life         
                                        organic               
                                        #food --------- #food 
                                                        flour 
                                                        yeast 
                                                        water 
                #bread -------------------------------- #bread
long ---------- long                                          
                colour ---- colour                             
                            white                             
                #colour --- #colour                            
crusty -------- crusty                                        
                #baguette - #baguette                         
                            #standard                         

(b)

                            special                           
                baguette -- baguette                          
                bread --------------------------------- bread 
                                        food ---------- food  
                                        ... ----------- ...   
sustains_life ------------------------- sustains_life         
                                        organic               
                                        #food --------- #food 
                                                        flour 
                                                        yeast 
                                                        water 
                #bread -------------------------------- #bread
long ---------- long                                          
                colour ---- colour                             
                            brown                             
                #colour --- #colour                            
crusty -------- crusty                                        
                #baguette - #baguette                         
                            rustic                            
                            #special                          

(c)
\end{BVerbatim}
\end{center}
\caption{The three best alignments formed by SP61 when the colour of a baguette is given a default value (`white') and Old contains a pattern for a special `rustic' class of baguette.}
\label{alignment_13}
\end{figure}

If we now add the symbol `rustic' to the symbols in New, only one alignment is formed that encodes all the symbols in New. This is shown in Figure \ref{alignment_14}. Notice how, with this additional information, the default colour for a baguette is overridden and we learn that, as a `rustic' baguette, the colour of the loaf is `brown'. And because there is only one alignment that encodes all the symbols in New, the relative probability of `brown' is 1.0.

\begin{figure}
\fontsize{08.00pt}{09.60pt}
\begin{center}
\begin{BVerbatim}
                            special                           
                baguette -- baguette                          
                bread --------------------------------- bread 
                                        food ---------- food  
                                        ... ----------- ...   
sustains_life ------------------------- sustains_life         
                                        organic               
                                        #food --------- #food 
                                                        flour 
                                                        yeast 
                                                        water 
                #bread -------------------------------- #bread
long ---------- long                                          
                colour ---- colour                             
                            brown                             
                #colour --- #colour                            
crusty -------- crusty                                        
                #baguette - #baguette                         
rustic -------------------- rustic                            
                            #special                          
\end{BVerbatim}
\end{center}
\caption{The best alignment formed by SP61 with `sustains\_life long crusty rustic' in New and the same patterns from Old as were used for Figure \ref{alignment_13}.}
\label{alignment_14}
\end{figure}

These examples illustrate the way in which inferences made by the system may be modified in the light of new information, in the manner of nonmonotonic reasoning.

\subsection{Other kinds of probabilistic inference}

Apart from the kinds of probabilistic reasoning described above, the ICMAUS framework lends itself to at least two other forms of inference where probabilities have a role:

\begin{itemize}

\item {\em Geometric Analogy Problems}. Given the translation of geometric forms into the kind of one-dimensional patterns currently required for SP61, it is possible to solve geometric analogy problems of the form `A is to B as C is to ?', with a choice of four possibilities for `?'. In essence, these kinds of problems can be understood as a search for good partial matches between patterns.

\item {\em Explaining Away `Explain Away'}. If we receive a phone call at work to say our burglar alarm has gone off, we are likely to assume there has a been a break in. However, if, at about the same time, we hear on the radio that there has been an earthquake, and if we know that the burglar alarm is sensitive to earthquakes, then we will probably assume that the earthquake is the explanation of why the alarm went off \citep{pearl_1988}. This kind of `explaining away' can be modelled by SP61 in a manner that is somewhat similar to the example of nonmonotonic reasoning, above. As with the previous example, the addition of new information (learning that an earthquake has occurred in this example) can change radically what kind of alignment of patterns achieves the most compression of the information in New.

\end{itemize}

Details of these examples may be found in \citet{wolff_1999_prob}.

\section{Unsupervised Inductive Learning}

As was noted earlier, this entire programme of research is based on earlier work on unsupervised inductive learning of language (Section \ref{background}), and the overall ICMAUS framework is designed to accommodate learning (Section \ref{icmaus_and_sp_models}). Most of the work to date has concentrated on areas other than learning but, as we noted in Section \ref{sp70_model}, the SP70 model has now been developed as a successor to SP61, designed as a full realisation of the ICMAUS framework, including unsupervised inductive learning. This section outlines how the SP70 model works. A relatively full description of the current version of SP70 is presented in \citet{wolff_unsupervised_learning}.

Imagine a new-born child listening to other people talking, either to the child directly or to each other. If ICMAUS principles apply, then each `New' portion of speech will be encoded as far as possible in terms of what is already stored. In accordance with empiricist thinking about language learning, we shall suppose that, initially, the child has no knowledge of language patterns. Thus each New portion of speech cannot be encoded in terms of anything else and is simply stored `as is'. After a time, however, as the child's brain gradually accumulates a collection of these speech patterns, new possibilities for economical encoding begin to appear.

Let us suppose that (the spoken equivalent of) `i t s b e d t i m e n o w' is already stored. Then, at some point, the child may hear something like `i t s p l a y t i m e n o w'. Looking for matching patterns, the child finds an alignment like the one shown at the top of Figure \ref{grammar_learning_figure_1}. By adding `code' symbols at appropriate points, the child may convert the alignment into a fragment of `grammar', something like the patterns shown in the bottom part of the figure. If patterns are stored in the brain as Hebbian cell assemblies, it is not too hard to imagine additional cells being added to each assembly to serve as identifiers or codes for the assembly.
\begin{figure}
\begin{center}
\begin{BVerbatim}
Alignment:

     0 i t s p l a y t i m e n o w 0
       | | |         | | | | | | |
     1 i t s b e d   t i m e n o w 1

Derived fragment of grammar:

     %1 i t s %2 #2 t i m e n o w #1
     %2 0 p l a y #2
     %2 1 b e d #2
\end{BVerbatim}
\end{center}
\caption{An alignment and corresponding grammar suggesting how a child may begin to identify words and classes of words in language.}
\label{grammar_learning_figure_1}
\end{figure}

Even at this early stage, the system has identified two discrete words and, in accordance with the principles of distributional linguistics, it has even recognised an embryonic grammatical class of `qualifying' words: {`play', `bed'}.

Notwithstanding nativist thinking about the nature of first language learning, there is now increasing recognition that distributional techniques can provide useful insights \citep{redington_chater_1998, redington_chater_finch_1998}. Demonstrations include my MK10 model of the learning of speech segmentation \citep{wolff_1975, wolff_1977, wolff_1980} and my SNPR model of syntax learning \citep{wolff_1982, wolff_1988, langley_stromsten_2000}.

\subsection{Features of SP70}\label{features_of_sp70}

This subsection indicates briefly some aspects of how the SP70 model works.

\subsubsection{Sifting and sorting}

Many of the alignments found by SP70 are not be as `tidy' as the example shown in Figure \ref{grammar_learning_figure_1}. It does not always happen that words or classes of words are picked out as cleanly as in the example. How can the system learn `clean' grammatical structures in the face of the inevitable messiness in the matching of linguistic patterns?

A related question is how children can learn `correct' grammars despite the fact that the language that they hear is often corrupted in various ways. That learning systems of the kind we have been discussing can cope with `dirty data' has been demonstrated already with the SNPR model of syntax learning.

The SNPR model can succeed in the face of dirty data because it is constantly searching for relatively large, relatively frequent patterns. Since `errors' in the raw data are, by their nature, relatively rare, they are sifted out and discarded in favour of the `good' patterns in the data.

SP70 incorporates a process for `sifting and sorting' through the patterns that have been derived from multiple alignments. This process compiles a set of alternative grammars, each one with a score based on MLE principles. The focus of interest is normally the best grammar produced by the program or the top two or three.

It is anticipated that, in future versions of the model, this sifting and sorting process will be applied at regular intervals as the system assimilates New information from its environment. At each stage, the store of Old pattern will be purged of all the patterns that are not part of the best grammar found by the system. In this way, the system may, in an incremental manner, build a store of Old patterns that are `good' in terms of MLE principles.

\subsubsection{Building hierarchical structures}

The toy example shown in Figure \ref{grammar_learning_figure_1} suggests how structures may be created at one level of abstraction above the raw data. Although SP70 can abstract plausible grammars from appropriate data, a weakness of the current version is that it is not good at finding levels of abstraction that are intermediate between the highest and lowest levels.

It is anticipated that this problem can be overcome by a reorganisation of the SP70 model. Incidentally, although the SNPR model lacks the generality of the ICMAUS framework, it does have the capability to discover structure at multiple levels of abstraction.

\subsubsection{Generalization of grammatical rules and correction of overgeneralizations}

Another important feature of the SNPR model is the way it can generalise grammatical rules and then correct over-generalisations despite the fact that, by definition, all kinds of generalisations, correct and incorrect, have zero frequency in the corpus from which the grammar is induced. This is achieved in a totally `unsupervised' way: without correction by a `teacher', without the provision of `negative' examples, and without any kind of `grading' of the material from which the system learns \citep{gold_1967}.

MLE principles appear to provide the key to distinguishing between `correct' and `incorrect' generalisations without external error correction: correct generalisations increase the compression that can be achieved while incorrect generalisations do the opposite.

SP70 already has an ability to make generalisations and to discriminate `good' ones from `bad' ones but further work is needed in this area.

\subsection{Unsupervised inductive learning of semantic structures}

The ICMAUS framework has been developed with the intention that it should be widely applicable, not confined narrowly to the syntax of natural language or some other circumscribed domain. One of the motivations for aiming for this kind of generality is that it should facilitate the learning of semantic knowledge structures and the integration of syntax with semantics.

Figure \ref{grammar_learning_figure_2} is intended to suggest how the kinds of class hierarchy considered in Section \ref{class_hierarchies} may be learned. The alignment at the top of the figure is intended to suggest how the features that are shared by swans and robins may be identified. These shared features (`wings feathers beak flies warm\_blooded') may be abstracted into a higher-level pattern describing the class `bird' (`\%bd wings feathers beak flies warm\_blooded \#bd') and then the two original patterns may be reduced to `\%sw swan \%bd \#bd long\_neck \#sw' and `\%rb robin \%bd \#bd red\_breast \#rb'.

\begin{figure}
\fontsize{09.00pt}{10.80pt}
\begin{center}
\begin{BVerbatim}
Alignment:

     0 swan  wings feathers beak flies warm_blooded long_neck  0
               |      |      |     |        |
     1 robin wings feathers beak flies warm_blooded red_breast 1

Derived fragment of grammar:

     %bd wings feathers beak flies warm_blooded #bd
     %sw swan %bd #bd long_neck #sw
     %rb robin %bd #bd red_breast #rb
\end{BVerbatim}
\end{center}
\caption{An alignment and corresponding grammar of non-linguistic patterns.}
\label{grammar_learning_figure_2}
\end{figure}

As with the learning of syntactic structures, it is anticipated that, when SP70 is more fully developed, this kind of thing will be done recursively so that arbitrarily deep hierarchical structures may be built up. At some stage, there will be a need to integrate this kind of learning with the learning of syntactic structures to achieve the kind of knowledge that can serve the process of understanding language and the process of producing language from meanings.

\section{Comparative Evaluation}\label{comparative_evaluation}

Although there is no space in this article to evaluate the ICMAUS framework properly in relation to alternative systems, a few remarks are offered here about how the ICMAUS system compares with Hidden Markov Models (HMMs) and Bayesian networks.

HMMs have been popular for applications like speech recognition but ``A major limitation [for speech] is the assumption that successive observations (frames of speech) are independent, and therefore the probability of a sequence of observations $P(O, O_2 ... O_T)$ can be written as a product of probabilities of individual observations ...'' \citep[][p. 284]{rabiner_1989}. Also, ``... the Markov assumption itself, i.e., that the probability of being in a given state at time $t$ only depends on the state at time $t - 1$, is clearly inappropriate for speech sounds.'' ({\em ibid.}).

These limitations of HMMs, which are true for kinds of knowledge other than speech, do not apply to the ICMAUS framework. Dependencies amongst observations are expressed in the ICMAUS framework by the use of patterns and, as we saw in Section \ref{syntactic_agreements}, these dependencies can bridge arbitrary amounts of intervening structure.

With appropriate design, HMMs have a memory for past events. But the same is true of the ICMAUS framework. In SP70, the repository of Old patterns is the cumulative memory of all the New patterns that have been received to date.

Bayesian networks have some attractive features for probabilistic reasoning and related applications \citep[see][]{pearl_1988} but they lack the flexibility of the ICMAUS framework for representing diverse kinds of knowledge. They are, for example, not well suited to representing the syntax of natural languages. By contrast, the ICMAUS framework provides a powerful means of handling such structures, including the subtle structure of interlocking dependencies in English auxiliary verbs \citep{wolff_2000}.

With regard to probabilities, each node in a Bayesian network must contain a table of transition probabilities and these can be quite complicated. By contrast, the ICMAUS framework stores all statistical information quite simply by attaching a frequency value to each stored pattern. This frequency information is used to calculate probabilities of particular contingencies as or when they are required.

\section{Conclusion}

This has been a relatively brief outline of the possibilities offered by the ICMAUS framework in understanding issues in cognitive science, artificial intelligence, and beyond. Despite the essential simplicity of the framework, it yields useful insights across a wide area.

A substantial programme of research has been needed to bring the ideas to their present stage of development. Now that they are relatively mature, I hope that other researchers will explore the several areas of application of the framework, will evaluate the framework and develop it further. As was noted earlier, the executable code and the source code for the SP61 model is now available and may be obtained from www.informatics.bangor.ac.uk/$\sim$gerry/sp\_summary.html. 

\section*{Acknowledgements}

I am very grateful for extensive and constructive comments on earlier drafts of this article that I have received from Dr Emmanuel Pothos of the Department of Psychology, University of Edinburgh and Professor Richard Young of the Department of Psychology, University of Hertfordshire. I am also very grateful to Mr John Hornsby and Dr Chris Whitaker of the School of Informatics, University of Wales Bangor, for advice on mathematical issues. The article has also benefited from constructive comments and suggestions by an anonymous reviewer.

This article, with revision and expansion, is based on ``Information compression by multiple alignment, unification and search as a framework for `intelligent' computing'' presented at the International ICSC Congress on Computational Intelligence: Methods \& Applications (CIMA 2001, June 2001, Bangor) and ``Information compression by multiple alignment, unification and search as a model of non-conscious intelligence'' presented at the Symposium on Nonconscious Intelligence: from Natural to Artificial (AISB'01 Convention, March 2001, York).


\begin{thebibliography}{48}
\expandafter\ifx\csname natexlab\endcsname\relax\def\natexlab#1{#1}\fi

\bibitem[Anderson and Lebiere(1998)]{anderson_lebiere_1998}
J.~R. Anderson and C.~J. Lebiere.
\newblock {\em The Atomic Components of Thought}.
\newblock Lawrence Erlbaum, Mahwah, NJ, 1998.

\bibitem[Antoniou(1997)]{antoniou_1997}
G.~Antoniou.
\newblock {\em Nonmonotonic Reasoning}.
\newblock MIT Press, Cambridge Mass., 1997.

\bibitem[Attneave(1954)]{attneave_1954}
F.~Attneave.
\newblock Some informational aspects of visual perception.
\newblock {\em Psychological Review}, 61:\penalty0 183--193, 1954.

\bibitem[Barlow(1969)]{barlow_1969}
H.~B. Barlow.
\newblock Trigger features, adaptation and economy of impulses.
\newblock In K.~N. Leibovic, editor, {\em Information Processes in the Nervous
  System}, pages 209--230. Springer, New York, 1969.

\bibitem[Barsalou(1992)]{barsalou_1992}
L.~W. Barsalou.
\newblock {\em Cognitive Psychology: An Overview for Cognitive Scientists}.
\newblock Lawrence Erlbaum, Hillsdale, NJ, 1992.

\bibitem[Chater(1996)]{chater_1996}
N.~Chater.
\newblock Reconciling simplicity and likelihood principles in perceptual
  organisation.
\newblock {\em Psychological Review}, 103\penalty0 (3):\penalty0 566--581,
  1996.

\bibitem[Chater(1999)]{chater_1999}
N.~Chater.
\newblock The search for simplicity: a fundamental cognitive principle?
\newblock {\em Quarterly Journal of Experimental Psychology}, 52 A\penalty0
  (2):\penalty0 273--302, 1999.

\bibitem[Collins and Quillian(1969)]{collins_quillian_1969}
A.~M. Collins and M.~R. Quillian.
\newblock Retrieval time from semantic memory.
\newblock {\em Journal of Verbal Learning and Verbal Behaviour}, 8:\penalty0
  240--248, 1969.

\bibitem[Cover and Thomas(1991)]{cover_thomas_1991}
T.~M. Cover and J.~A. Thomas.
\newblock {\em Elements of Information Theory}.
\newblock John Wiley, New York, 1991.

\bibitem[Garner(1974)]{garner_1974}
W.~R. Garner, editor.
\newblock {\em The Processing of Information and Structure}.
\newblock Lawrence Erlbaum, Hillsdale, NJ, 1974.

\bibitem[Gazdar and Mellish(1989)]{gazdar_mellish_1989}
G.~Gazdar and C.~Mellish.
\newblock {\em Natural Language Processing in Prolog}.
\newblock Addison-Wesley, Wokingham, 1989.

\bibitem[Ginsberg(1994)]{ginsberg_1994}
M.~L. Ginsberg.
\newblock {AI} and nonmonotonic reasoning.
\newblock In D.~M. Gabbay, C.~J. Hogger, and J.~A. Robinson, editors, {\em
  Handbook of Logic in Artificial Intelligence and Logic Programming:
  Nonmonotonic Reasoning and Uncertain Reasoning}, volume~3, pages 1--33.
  Oxford University Press, Oxford, 1994.

\bibitem[Gold(1967)]{gold_1967}
M.~Gold.
\newblock Language identification in the limit.
\newblock {\em Information and Control}, 10:\penalty0 447--474, 1967.

\bibitem[Laird et~al.(1987)Laird, Newell, and
  Rosenbloom]{laird_newell_rosenbloom_1987}
J.~E. Laird, A.~Newell, and P.~S. Rosenbloom.
\newblock Soar: an architecture for general intelligence.
\newblock {\em Artificial Intelligence}, 33:\penalty0 1--64, 1987.

\bibitem[Langley and Stromsten(2000)]{langley_stromsten_2000}
P.~Langley and S.~Stromsten.
\newblock Learning context-free grammars with a simplicity bias.
\newblock In {\em Proceedings of the Eleventh European Conference on Machine
  Learning}, pages 220--228, 2000.
\newblock Barcelona.

\bibitem[Li and Vit\'{a}nyi(1997)]{li_vitanyi_1997}
M.~Li and P.~Vit\'{a}nyi.
\newblock {\em An Introduction to Kolmogorov Complexity and Its Applications}.
\newblock Springer-Verlag, New York, 1997.

\bibitem[Minsky(1967)]{minsky_1967}
M.~L. Minsky.
\newblock {\em Computation, Finite and Infinite Machines}.
\newblock Prentice Hall, Englewood Cliffs, NJ., 1967.

\bibitem[Newell(1992)]{newell_1992}
A.~Newell.
\newblock Pr\'{e}cis of ``unified theories of cognition''.
\newblock {\em Behavioural and Brain Sciences}, 15:\penalty0 425--492, 1992.

\bibitem[Oldfield(1954)]{oldfield_1954}
R.~C. Oldfield.
\newblock Memory mechanisms and the theory of schemata.
\newblock {\em British Journal of Psychology}, 45:\penalty0 14--23, 1954.

\bibitem[Pearl(1988)]{pearl_1988}
J.~Pearl.
\newblock {\em Probabilistic Reasoning in Intelligent Systems}.
\newblock Morgan Kaufmann, San Francisco, 1988.

\bibitem[Pereira and Warren(1980)]{pereira_warren_1980}
F.~C.~N. Pereira and D.~H.~D. Warren.
\newblock Definite clause grammars for language analysis - a survey of the
  formalism and a comparison with augmented transition networks.
\newblock {\em Artificial Intelligence}, 13:\penalty0 231--278, 1980.

\bibitem[Rabiner(1989)]{rabiner_1989}
L.~R. Rabiner.
\newblock A tutorial on hidden markov models and selected applications in
  speech recognition.
\newblock {\em Proceedings of the IEEE}, 77\penalty0 (2):\penalty0 257--286,
  1989.

\bibitem[Redington and Chater(1998)]{redington_chater_1998}
M.~Redington and N.~Chater.
\newblock Connectionist and statistical approaches to language acquisition: a
  distributional perspective.
\newblock {\em Language and Cognitive Processes}, 1\penalty0 (2/3):\penalty0
  129--191, 1998.

\bibitem[Redington et~al.(1998)Redington, Chater, and
  Finch]{redington_chater_finch_1998}
M.~Redington, N.~Chater, and S.~Finch.
\newblock Distributional information: a powerful cue for acquiring syntactic
  categories.
\newblock {\em Cognitive Science}, 22\penalty0 (4):\penalty0 425--469, 1998.

\bibitem[Rissanen(1978)]{rissanen_1978}
J.~Rissanen.
\newblock Modelling by the shortest data description.
\newblock {\em Automatica-J, {IFAC}}, 14:\penalty0 465--471, 1978.

\bibitem[Sankoff and Kruskall(1983)]{sankoff_kruskall_1983}
D.~Sankoff and J.~B. Kruskall.
\newblock {\em Time Warps, String Edits, and Macromolecules: the Theory and
  Practice of Sequence Comparisons}.
\newblock Addison-Wesley, Reading, MA, 1983.

\bibitem[Solomonoff(1964)]{solomonoff_1964}
R.~J. Solomonoff.
\newblock A formal theory of inductive inference. parts {I} and {II}.
\newblock {\em Information and Control}, 7:\penalty0 1--22 and 224--254, 1964.

\bibitem[Solomonoff(1986)]{solomonoff_1986}
R.~J. Solomonoff.
\newblock The application of algorithmic probability to problems in artificial
  intelligence.
\newblock In L.~N. Kanal and J.~F. Lemmer, editors, {\em Uncertainty in
  Artificial Intelligence}, pages 473--491. Elsevier Science, North-Holland,
  1986.

\bibitem[Solomonoff(1997)]{solomonoff_1997}
R.~J. Solomonoff.
\newblock The discovery of algorithmic probability.
\newblock {\em Journal of Computer and System Sciences}, 55\penalty0
  (1):\penalty0 73--88, 1997.

\bibitem[{von B{\'e}k{\'e}sy}(1967)]{von_bekesy_1967}
G.~{von B{\'e}k{\'e}sy}.
\newblock {\em Sensory Inhibition}.
\newblock Princeton University Press, Princeton, NJ, 1967.

\bibitem[Wallace and Boulton(1968)]{wallace_boulton_1968}
C.~S. Wallace and D.~M. Boulton.
\newblock An information measure for classification.
\newblock {\em Computer Journal}, 11\penalty0 (2):\penalty0 185--195, 1968.

\bibitem[Watanabe(1972)]{watanabe_article_1972}
S.~Watanabe.
\newblock Pattern recognition as information compression.
\newblock In S.~Watanabe, editor, {\em Frontiers of Pattern Recognition}.
  Academic Press, New York, 1972.

\bibitem[Wolff(1975)]{wolff_1975}
J.~G. Wolff.
\newblock An algorithm for the segmentation of an artificial language analogue.
\newblock {\em British Journal of Psychology}, 66:\penalty0 79--90, 1975.

\bibitem[Wolff(1977)]{wolff_1977}
J.~G. Wolff.
\newblock The discovery of segments in natural language.
\newblock {\em British Journal of Psychology}, 68:\penalty0 97--106, 1977.

\bibitem[Wolff(1980)]{wolff_1980}
J.~G. Wolff.
\newblock Language acquisition and the discovery of phrase structure.
\newblock {\em Language \& Speech}, 23:\penalty0 255--269, 1980.

\bibitem[Wolff(1982)]{wolff_1982}
J.~G. Wolff.
\newblock Language acquisition, data compression and generalization.
\newblock {\em Language \& Communication}, 2:\penalty0 57--89, 1982.

\bibitem[Wolff(1988)]{wolff_1988}
J.~G. Wolff.
\newblock Learning syntax and meanings through optimization and distributional
  analysis.
\newblock In Y.~Levy, I.~M. Schlesinger, and M.~D.~S. Braine, editors, {\em
  Categories and Processes in Language Acquisition}, pages 179--215. Lawrence
  Erlbaum, Hillsdale, NJ, 1988.

\bibitem[Wolff(1990)]{wolff_1990}
J.~G. Wolff.
\newblock Simplicity and power---some unifying ideas in computing.
\newblock {\em Computer Journal}, 33\penalty0 (6):\penalty0 518--534, 1990.

\bibitem[Wolff(1993)]{wolff_1993}
J.~G. Wolff.
\newblock Computing, cognition and information compression.
\newblock {\em AI Communications}, 6\penalty0 (2):\penalty0 107--127, 1993.

\bibitem[Wolff(1994)]{wolff_1994_scaleable}
J.~G. Wolff.
\newblock A scaleable technique for best-match retrieval of sequential
  information using metrics-guided search.
\newblock {\em Journal of Information Science}, 20\penalty0 (1):\penalty0
  16--28, 1994.

\bibitem[Wolff(1999{\natexlab{a}})]{wolff_1999_comp}
J.~G. Wolff.
\newblock {`Computing'} as information compression by multiple alignment,
  unification and search.
\newblock {\em Journal of Universal Computer Science}, 5\penalty0
  (11):\penalty0 777--815, 1999{\natexlab{a}}.

\bibitem[Wolff(1999{\natexlab{b}})]{wolff_1999_prob}
J.~G. Wolff.
\newblock Probabilistic reasoning as information compression by multiple
  alignment, unification and search: an introduction and overview.
\newblock {\em Journal of Universal Computer Science}, 5\penalty0 (7):\penalty0
  418--462, 1999{\natexlab{b}}.

\bibitem[Wolff(2000)]{wolff_2000}
J.~G. Wolff.
\newblock Syntax, parsing and production of natural language in a framework of
  information compression by multiple alignment, unification and search.
\newblock {\em Journal of Universal Computer Science}, 6\penalty0 (8):\penalty0
  781--829, 2000.

\bibitem[Wolff(2002{\natexlab{a}})]{wolff_maths_logic}
J.~G. Wolff.
\newblock Mathematics and logic as information compression by multiple
  alignment, unification and search.
\newblock Technical report, CognitionResearch.org.uk, 2002{\natexlab{a}}.
\newblock Copy: www.cognitionresearch.org.uk/papers/cml/cml.htm.

\bibitem[Wolff(2002{\natexlab{b}})]{wolff_icmaus_neural}
J.~G. Wolff.
\newblock Neural mechanisms for information compression by multiple alignment,
  unification and search.
\newblock Technical report, CognitionResearch.org.uk, 2002{\natexlab{b}}.
\newblock Available from the author on request.

\bibitem[Wolff(2002{\natexlab{c}})]{wolff_unsupervised_learning}
J.~G. Wolff.
\newblock Unsupervised learning in a framework of information compression by
  multiple alignment, unification and search.
\newblock Technical report, CognitionResearch.org.uk, 2002{\natexlab{c}}.
\newblock Copy: http://uk.arxiv.org/abs/cs.AI/0302015.

\bibitem[Young and Lewis(1999)]{young_lewis_1999}
R.~M. Young and R.~L. Lewis.
\newblock The {S}oar cognitive architecture and human working memory.
\newblock In A.~Miyake and P.~Shah, editors, {\em Models of Working Memory:
  Mechanisms of Active Maintenance and Executive Control}, pages 224--256.
  Cambridge University Press, Cambridge, 1999.
\newblock Chapter 7.

\bibitem[Zipf(1949)]{zipf_1949}
G.~K. Zipf.
\newblock {\em Human Behaviour and the Principle of Least Effort}.
\newblock Hafner, New York, 1949.

\end{thebibliography}
\end{document}